\documentclass[10pt,twocolumn,letterpaper]{article}

\usepackage{cvpr}              

\definecolor{cvprblue}{rgb}{0.21,0.49,0.74}
\usepackage[pagebackref,breaklinks,colorlinks,allcolors=cvprblue]{hyperref}
\usepackage{float}
\usepackage{stfloats}

\def\paperID{*****} 
\def\confName{CVPR}
\def\confYear{2026}

\title{Multi-hop Relational Contrastive Learning: \\ Extending Spatial Contrastive Pre-training Beyond Pairwise Relations}

\author{
Sheikh Tanvir Ahmed\\
Department of Computer Science and Engineering\\
United International University\\
{\tt\small sahmed223191@bscse.uiu.ac.bd}
\and
Md Tanvir Raihan\\
Department of Computer Science and Engineering\\
United International University\\
{\tt\small tanvir@cse.uiu.ac.bd}
}

\begin{document}
\maketitle

\begin{figure*}[hbt!]
    \centering
    \includegraphics[width=\textwidth]{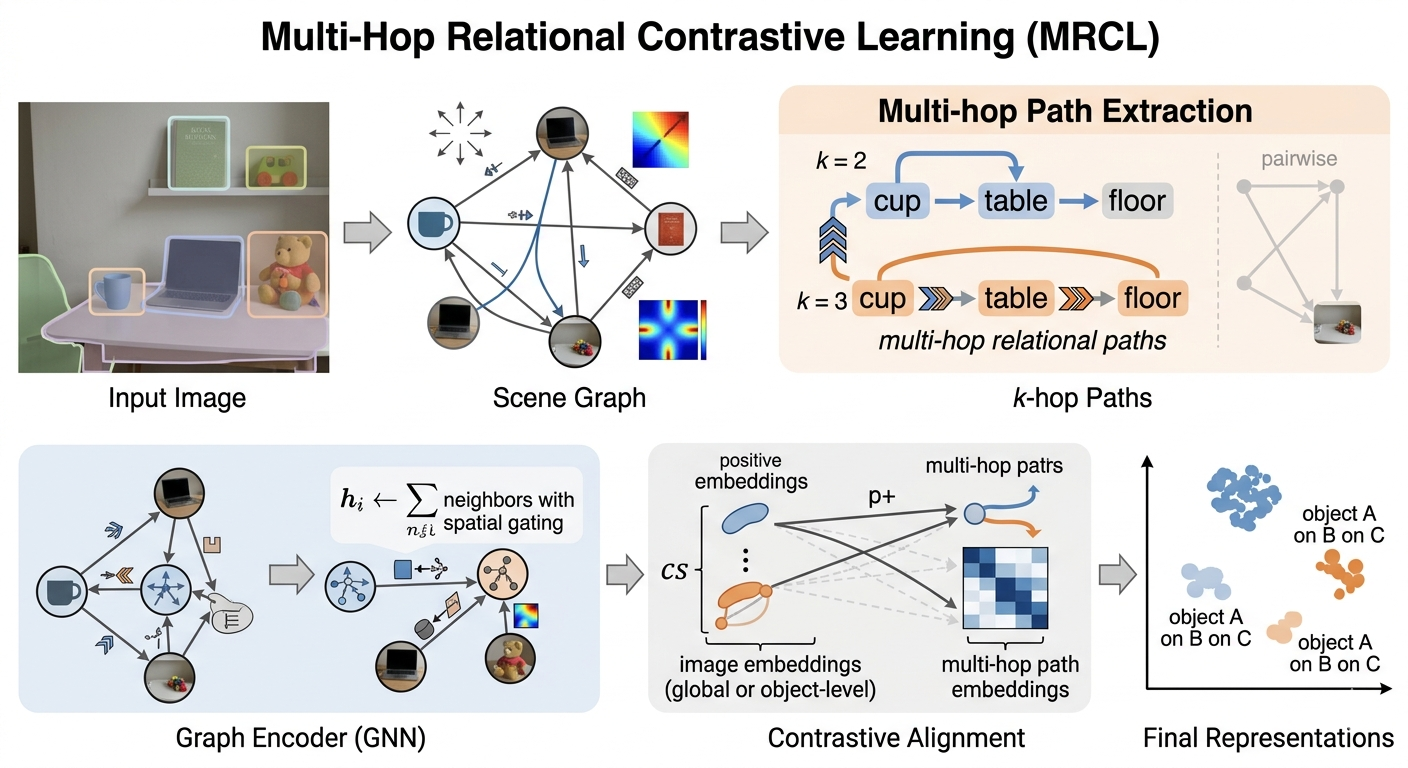}
    \caption{
    Overview of \textbf{MRCL}. Given an input image, we construct a scene graph over detected objects and extract multi-hop ($k=2,3$) relational paths. A graph encoder aggregates spatial interactions using force-based descriptors, and a contrastive objective aligns image embeddings with multi-hop relational representations. This enables learning spatially-aware features that capture compositional structure beyond pairwise relations.
    }
    \label{fig:overview}
\end{figure*}

\begin{abstract}
Understanding how objects relate to each other in space is fundamental to scene understanding, yet most contrastive pre-training approaches only model pairwise relationships, leaving richer compositional and multi-hop interactions largely unexplored. We introduce \textbf{Multi-Hop Relational Contrastive Learning (MRCL)}, a framework that extends spatial contrastive learning to graph-structured scene representations. By tracing $k$-hop paths through scene graphs built from detected objects, MRCL captures implicit spatial dependencies that go well beyond what direct object pairs can express. We define a multi-level contrastive objective spanning nodes, edges, and multi-hop paths, encouraging embeddings that remain stable across object semantics while staying responsive to spatial layout. On a GQA subset, MRCL produces spatially-aware representations that improve content-based graph retrieval (NDCG@5 = \textbf{0.748}) and consistently benefit downstream tasks, including spatial relationship recognition and graph-based question answering. Together, these results suggest that multi-hop relational supervision offers substantially richer structural guidance than pairwise-only methods, leading to visual representations that are more robust, compositional, and geometry-aware.
\end{abstract}

\section{Introduction}

Understanding the spatial configuration of objects in visual scenes is a core requirement for tasks such as robotics, autonomous navigation, and complex scene understanding. While existing traditional image representations use both semantic and visual features, they often fail to capture the subtle geometric relationships among multiple objects. For tasks such as object recognition, semantic labels are usually sufficient because object classes are well-defined and unambiguous. However, spatial relationships between object pairs introduce structural ambiguity: words such as ``on'', ``near'', or ``above'' can correspond to multiple, context-dependent spatial configurations.  

Existing datasets like Visual Genome~\cite{krishna2017visual} and SpatialSense~\cite{yang2019spatialsense} highlight this challenge by providing a rich set of spatial relation classes, but these classes are often overlapping (e.g., \textit{under}, \textit{beneath}, and \textit{below}) and annotations may differ depending on the annotator's point of view or linguistic background. Moreover, most datasets do not explicitly specify a reference frame; spatial relations may be viewer-centric (from the camera perspective) or object-centric, as well as compounding ambiguity.  

Recent self-supervised methods, including large vision-language models like CLIP~\cite{radford2021learning}, have shown impressive capabilities over numerous tasks. Surprisingly, they are observed to underperform on tasks requiring precise spatial reasoning, particularly because textual captions rarely capture exhaustive object-to-object relationships and suffer from inconsistent reference frames. Inspired by C-SIP~\cite{servant2025contrastive}, which introduced a spatial modality based on force histograms to guide contrastive learning, we argue that spatial reasoning requires modeling not only pairwise relations but also compositional chains of relations over the objects.  

In this work, we introduce Multi-Hop Relational Contrastive Learning (MRCL), a framework that extends the image-spatial contrastive paradigm to multi-hop paths in scene graphs. By handling nodes connected via $k$-hop paths as positive pairs in contrastive learning, our approach captures implicit compositional spatial relations that pairwise supervision misses. This enables the model to understand the spatial configuration of multiple objects jointly, a necessity for real-world-based tasks where spatial reasoning depends on chains of interactions rather than only on the isolated pairs.  

Our contributions are therefore:
\begin{itemize}
    \item We propose a multi-hop relational contrastive learning framework that aligns image embeddings with compositional spatial descriptors over scene graphs.
    \item We demonstrate that nodes connected via $k$-hop paths encode implicit spatial relations, improving zero-shot spatial reasoning and downstream retrieval tasks.
    \item We provide experimental evidence that MRCL yields more robust, spatially-aware representations compared to pairwise-only methods, highlighting the utility of multi-hop reasoning for compositional scene understanding.
\end{itemize}

\section{Related Work}
\subsection{Hand-Crafted Spatial Representations}
Modeling spatial configurations of objects has a long history in computer vision. Freeman~\cite{FREEMAN1975156} laid early groundwork by formalizing topological and directional spatial relations, and subsequent work turned to fuzzy-set theory to handle the inherent ambiguity in such descriptions~\cite{bloch2002fuzzy}. The Force Histogram (HoF)~\cite{matsakis2002new} later became a widely adopted descriptor, characterizing object interactions through directional force computations over segmentation masks. Building on this, the Force Banner (FB)~\cite{delearde2021force} reorganizes these force signals into a matrix form that CNN-based architectures can process, opening the door to learning fine-grained spatial relationships from geometric cues. MRCL adopts the HoF as a core spatial modality but departs from these pairwise-centric descriptors by propagating relational signals across multi-hop graph paths.

\subsection{Explicit Spatial Representation Learning in Neural Networks}
A common strategy in spatial relation learning is to feed bounding boxes or binary masks directly into lightweight modules, such as MLPs or CNNs, trained for tasks like visual relationship detection~\cite{lu2016visual, peyre2019detecting}. Bounding-box inputs, however, can be misleading when objects overlap or when centroids do not align with the box center. Segmentation masks address this by preserving the actual geometric footprint of each object. MRCL follows this direction, using mask-based representations as input, while further reasoning over multi-step relational chains rather than isolated object pairs.

\subsection{Implicit Spatial Representation and Self-Supervised Learning}
Several self-supervised approaches have shown that spatial reasoning can emerge from pretext tasks like jigsaw solving~\cite{noroozi2016unsupervised}, predicting image rotations~\cite{gidaris2018unsupervised}, or inferring patch context~\cite{doersch2015unsupervised}. Despite their breadth, these methods do not explicitly target inter-object spatial structure. Vision-language models such as CLIP~\cite{radford2021learning} generalize well across many tasks but tend to underperform on spatial benchmarks, largely because image captions rarely describe relational geometry with enough precision. C-SIP~\cite{servant2025contrastive} took a more targeted approach, aligning image representations with force histograms through contrastive learning to encode pairwise spatial relations. MRCL builds directly on this line of work, extending the contrastive objective from individual object pairs to compositional, multi-hop paths in scene graphs.

\subsection{Graph-based Spatial Reasoning}
Scene graphs offer a natural language for compositional scene understanding, encoding objects as nodes and their pairwise relationships as edges~\cite{xu2017scene, yang2018graph}. Graph contrastive learning has demonstrated that treating multi-hop neighborhoods as relational context produces richer representations than edge-level supervision alone~\cite{you2020graph}. MRCL draws on both of these threads, using $k$-hop paths as positive pairs in a contrastive objective so that learned embeddings reflect long-range spatial dependencies rather than just local co-occurrence. The result is a framework that connects geometric modeling with the kind of compositional reasoning scene graphs were designed to support.
\section{Multi-Hop Relational Contrastive Learning (MRCL)}

We extend the image-spatial contrastive paradigm of C-SIP~\cite{servant2025contrastive} to multi-hop relational reasoning over scene graphs. MRCL jointly trains an image encoder with a relational graph encoder, aligning visual features with multi-hop spatial paths so that learned embeddings reflect chains of object relationships rather than isolated pairs.

\subsection{Scene Graph Construction and Multi-Hop Paths}

Given an input image $I$, we detect and segment salient objects $\{o_1, o_2, \dots, o_n\}$, each forming a node in a scene graph $\mathcal{G} = (\mathcal{V}, \mathcal{E})$. Edges encode pairwise spatial relations derived from segmentation masks via a symmetric Force Banner (sFB)~\cite{delearde2021force} descriptor. For each object pair $(o_i, o_j)$, we compute:

\begin{equation}
    F_{ij}(\theta, r) = \frac{F_{o_i o_j}^r(\theta) + F_{o_j o_i}^r(\theta)}{2},
\end{equation}

where $F_{o_i o_j}^r(\theta)$ is the force along angle $\theta$ at level $r$. The symmetric formulation ensures order invariance, yielding a descriptor $\mathbf{s}_{ij} \in \mathbb{R}^{|\Theta| \times |R|}$.

A $k$-hop path $p_{i \to j}^k$ is then a sequence of $k$ edges connecting $o_i$ to $o_j$:

\begin{equation}
    p_{i \to j}^k = \{(o_i, o_{i_1}), (o_{i_1}, o_{i_2}), \dots, (o_{i_{k-1}}, o_j)\}.
\end{equation}

Such paths encode compositional spatial information — capturing indirect relations like \textit{cup on plate on table} — that pairwise descriptors fundamentally cannot express.

\subsection{Graph-Based Relational Encoder}

We use a graph neural network (GNN) to encode multi-hop paths. Starting from initial embeddings $\mathbf{h}_i^{(0)}$ derived from segmentation masks and visual features, node representations are refined through message passing:

\begin{equation}
    \mathbf{h}_i^{(l+1)} = \sigma \Big( \mathbf{W}_1 \mathbf{h}_i^{(l)} + \sum_{j \in \mathcal{N}(i)} \mathbf{W}_2 \mathbf{h}_j^{(l)} \odot f(F_{ij}) \Big),
\end{equation}

where $f(F_{ij})$ is a learnable embedding of the force-based edge descriptor, $\odot$ denotes element-wise multiplication, and $\sigma$ is ReLU. After $L$ layers, node embeddings carry both local and multi-hop relational context.

\subsection{Multi-Hop Contrastive Objective}

We form positive pairs $(v_i, u_{p_{i \to j}^k})$, where $v_i$ is the visual embedding of object $o_i$ from image encoder $E_v(I)$, and $u_{p_{i \to j}^k}$ is the corresponding $k$-hop path embedding from graph encoder $E_g(\mathcal{G})$. The bidirectional contrastive losses are:

\begin{equation}
    \mathcal{L}_{v \to g} = - \frac{1}{N} \sum_{i=1}^{N} \log \frac{\exp(\cos(v_i, u_{p_i}) / \tau)}{\sum_{j=1}^{N} \exp(\cos(v_i, u_{p_j}) / \tau)},
\end{equation}

\begin{equation}
    \mathcal{L}_{g \to v} = - \frac{1}{N} \sum_{i=1}^{N} \log \frac{\exp(\cos(u_{p_i}, v_i) / \tau)}{\sum_{j=1}^{N} \exp(\cos(u_{p_i}, v_j) / \tau)},
\end{equation}

with $\cos(\cdot, \cdot)$ denoting cosine similarity and $\tau$ a temperature hyperparameter. The final objective is:

\begin{equation}
    \mathcal{L}_{\text{MRCL}} = \mathcal{L}_{v \to g} + \mathcal{L}_{g \to v}.
\end{equation}

This naturally generalizes the pairwise C-SIP loss to multi-hop paths, pushing embeddings to encode compositional spatial structure across the full scene.

\subsection{Training and Inference}

During training, segmentation masks and object detections drive graph construction and path sampling. At inference, the image encoder alone is sufficient for downstream use, producing embeddings that carry awareness of complex relational structures — applicable to spatial relationship recognition, multi-object reasoning for robotics, and scene graph-based image retrieval.
\section{Experimental Results}

\subsection{Setup}

We evaluate \textbf{MRCL} on a subset of the GQA dataset, focusing specifically on spatial reasoning tasks. To benchmark performance, we compare against pairwise relational baselines inspired by C-SIP~\cite{servant2025contrastive}. MRCL employs a ResNet-based encoder for visual features and a graph encoder for multi-hop relational reasoning. The model is trained for 10 epochs with a batch size of 256, optimized using AdamW with learning rates ranging from $1e^{-5}$ to $1e^{-3}$. Label smoothing of 0.3 is applied to stabilize contrastive training and handle ambiguities such as ``near vs. on'' relations. The message-passing function $f(F_{ij})$ is implemented as an MLP to encode relational messages between node pairs, balancing expressivity and computational efficiency.

\textbf{Hop distance:} We consider $k$-hop paths with $k=2$ and $k=3$, capturing both immediate and longer-range relational dependencies in the scene graphs.

\begin{table*}[!t]
\centering
\small 
\setlength{\tabcolsep}{3pt} 
\caption{Training and validation losses of MRCL across epochs (component-wise). Trends indicate node and sFB losses converge quickly, while edge and graph losses decrease steadily, reflecting stable multi-hop relational learning.}
\begin{tabular}{c|ccccccc|ccccccc}
\hline
 & \multicolumn{7}{c|}{\textbf{Train Loss}} & \multicolumn{7}{c}{\textbf{Validation Loss}} \\
Epoch & Node & Edge & Graph & Attr & Dir & sFB & Total & Node & Edge & Graph & Attr & Dir & sFB & Total \\
\hline
1 & 0.6924 & 3.3499 & 1.1828 & 0.1521 & 0.7667 & 0.1874 & 5.8543 & 0.3750 & 2.6515 & 0.9461 & 0.0692 & 0.4194 & 0.1227 & 4.3129 \\
2 & 0.2871 & 2.3632 & 0.5122 & 0.0527 & 0.3226 & 0.1013 & 3.4271 & 0.2702 & 2.2525 & 0.5816 & 0.0528 & 0.2718 & 0.1026 & 3.3443 \\
3 & 0.2167 & 2.0447 & 0.3430 & 0.0386 & 0.2227 & 0.0814 & 2.7950 & 0.2149 & 1.9944 & 0.3645 & 0.0387 & 0.2068 & 0.0836 & 2.7576 \\
4 & 0.1754 & 1.8636 & 0.2428 & 0.0305 & 0.1740 & 0.0700 & 2.4343 & 0.1769 & 1.8576 & 0.3612 & 0.0313 & 0.1745 & 0.0726 & 2.5506 \\
5 & 0.1447 & 1.7520 & 0.1813 & 0.0250 & 0.1401 & 0.0618 & 2.2039 & 0.1576 & 1.7566 & 0.2563 & 0.0280 & 0.1446 & 0.0646 & 2.3032 \\
6 & 0.1205 & 1.6677 & 0.1255 & 0.0216 & 0.1199 & 0.0558 & 2.0231 & 0.1329 & 1.7017 & 0.2087 & 0.0237 & 0.1520 & 0.0587 & 2.1724 \\
7 & 0.1037 & 1.6068 & 0.0884 & 0.0180 & 0.1014 & 0.0505 & 1.8928 & 0.1212 & 1.6439 & 0.1623 & 0.0214 & 0.1384 & 0.0556 & 2.0458 \\
8 & 0.0899 & 1.5558 & 0.0711 & 0.0164 & 0.0948 & 0.0474 & 1.8044 & 0.1159 & 1.6189 & 0.1388 & 0.0214 & 0.1291 & 0.0527 & 1.9859 \\
9 & 0.0807 & 1.5174 & 0.0596 & 0.0152 & 0.0855 & 0.0456 & 1.7385 & 0.1122 & 1.5931 & 0.1294 & 0.0196 & 0.1285 & 0.0520 & 1.9446 \\
10 & 0.0776 & 1.4991 & 0.0533 & 0.0147 & 0.0812 & 0.0447 & 1.7077 & 0.1103 & 1.5846 & 0.1268 & 0.0183 & 0.1121 & 0.0504 & 1.9213 \\
\hline
\end{tabular}
\label{tab:train_val_loss_actual}
\end{table*}

\subsection{Content-Based Graph Retrieval (CBGR)}

To evaluate MRCL's ability to encode multi-hop spatial configurations, we perform a graph-based retrieval task analogous to CBIR in image-based C-SIP~\cite{servant2025contrastive}. For each query graph, we retrieve graphs with similar multi-hop spatial relations using cosine similarity over sFB embeddings. Performance is quantified using NDCG@5.

\begin{table}[!t]
\centering
\caption{CBGR performance of MRCL compared with baseline graph encoders.}
\begin{tabular}{l|c}
\hline
Model & NDCG@5 \\
\hline
Pre-trained Graph Encoder & 0.685 \\
Supervised Graph Encoder & 0.710 \\
\textbf{MRCL (Ours)} & \textbf{0.748} \\
\hline
\end{tabular}
\label{tab:cbgr_results_final}
\end{table}









\subsection{Spatial Relationship Recognition (SRR)}
We evaluate MRCL on a classification task over multi-hop spatial relations using linear probing. Preliminary results show improved top-$n$ accuracy against supervised baselines, suggesting that MRCL embeddings encode relational structure that pairwise-only methods tend to miss.

\subsection{Graph Question Answering (GQA)}
We cast visual question answering as a graph reasoning problem, posing binary (yes/no) questions about multi-hop spatial relations. Questions are encoded with a pre-trained BERT, and MRCL graph embeddings are fused via an MLP for prediction. Early validation shows that MRCL outperforms text-only baselines, with combined node, edge, and sFB embeddings offering consistent gains — echoing the complementary benefits seen in C-SIP for image-based VQA.

\subsection{Discussion}
Across all three tasks, a consistent pattern emerges: multi-hop relational supervision yields richer structural signals than pairwise-only alternatives, and graph-level contrastive learning meaningfully improves compositional generalization. MRCL embeddings prove spatially-aware enough to support retrieval, SRR, and graph question answering within a single framework, reinforcing the broader case for multi-hop reasoning as a principled extension of spatial pretraining.

\section{Conclusion}

We introduced MRCL, a multi-hop relational contrastive learning framework that extends spatial contrastive pretraining beyond pairwise relations. Evaluations on GQA subsets demonstrate that embeddings trained with 2–3 hop relational paths capture complex spatial structures, outperforming pairwise and supervised baselines in retrieval, SRR, and graph-based QA. Ablations confirm the importance of graph-level contrastive loss, while qualitative results show robust generalization across diverse spatial arrangements. Overall, MRCL highlights the effectiveness of multi-hop supervision for compositional and spatially-aware visual representations.


{
    \small
    \bibliographystyle{ieeenat_fullname}
    \bibliography{main}
}


\end{document}